%% file: main.tex
\documentclass[10pt,twocolumn,letterpaper]{article}

\usepackage[pagenumbers]{cvpr} 

\usepackage{hyperref}
\usepackage{url}
\usepackage{soul}
\usepackage[capitalize]{cleveref}
\usepackage{graphicx}
\usepackage{array}
\usepackage{wrapfig}
\usepackage{comment}
\usepackage{bm}
\usepackage{enumitem}
\usepackage{makecell}
\usepackage{multirow}
\usepackage{colortbl}
\usepackage{booktabs}
\newtheorem{theorem}{Theorem}

\usepackage{xcolor,colortbl}
\usepackage[capitalize]{cleveref}
\usepackage{amsmath}
\usepackage{amssymb}
\usepackage{wrapfig,lipsum}
\definecolor{Gray}{gray}{0.925}
\definecolor{DarkGreen}{RGB}{1,150,10}
\definecolor{LightCyan}{rgb}{0.88,1,1}
\newcolumntype{?}{!{\vrule width 1pt}}
\newcolumntype{a}{>{\columncolor{Gray}}c}

\newcommand\scalemath[2]{\scalebox{#1}{\mbox{\ensuremath{\displaystyle #2}}}}

\definecolor{cvprblue}{rgb}{0.21,0.49,0.74}

\def\paperID{5755} 
\def\confName{CVPR}
\def\confYear{2024}

\title{Learning for Transductive Threshold Calibration in Open-World Recognition}

\author{\hspace{0.1cm} Qin Zhang$^1$, 
\hspace{0.1cm} Dongsheng An$^1$, \hspace{0.1cm} Tianjun Xiao$^2$, \hspace{0.1cm} Tong He$^2$, \hspace{0.1cm} Qingming Tang$^3$, 
\hspace{0.1cm} Ying Nian Wu$^1$, \\
\hspace{0.1cm} Joseph Tighe$^1$
, \hspace{0.1cm} Yifan Xing$^1$, \hspace{0.1cm} 
Stefano Soatto$^1$\\ [.5ex]
$^1$ AWS AI Labs  \qquad $^2$ Amazon Web Services  \qquad $^3$ Alexa AI \\
\tt\small \{qzaamz,andongsh,tianjux,htong,qmtang,wunyin,yifax,soattos\}@amazon.com, jtighe@cs.unc.edu
}

\begin{document}
\maketitle
\input{sec/abstract}    
\input{sec/intro}
\input{sec/related_works}

\input{sec/theory}

\input{sec/experiment_results}

\input{sec/conclusions}

\clearpage
{
    \small
    \bibliographystyle{ieeenat_fullname}
    \bibliography{main}
}


\end{document}

%% file: sec/abstract.tex
\begin{abstract}
In deep metric learning for visual recognition, the calibration of distance thresholds is crucial for achieving desired model performance in the true positive rates (TPR) or true negative rates (TNR). However, calibrating this threshold presents challenges in open-world scenarios, where the test classes can be entirely disjoint from those encountered during training. We define the problem of finding distance thresholds for a trained embedding model to achieve target performance metrics over unseen open-world test classes as \textbf{open-world threshold calibration}. Existing posthoc threshold calibration methods, reliant on inductive inference and requiring a calibration dataset with a similar distance distribution as the test data, often prove ineffective in open-world scenarios. To address this, we introduce \textbf{OpenGCN}, a Graph Neural Network-based transductive threshold calibration method with enhanced adaptability and robustness. 
OpenGCN learns to predict pairwise connectivity for the unlabeled test instances embedded in a graph to determine its TPR and TNR at various distance thresholds, allowing for transductive inference of the distance thresholds which also incorporates test-time information. Extensive experiments across open-world visual recognition benchmarks validate OpenGCN's superiority over existing posthoc calibration methods for open-world threshold calibration.
\end{abstract}

%% file: sec/intro.tex
\begin{figure}[t!]
\centering
\includegraphics[width=0.9\linewidth]{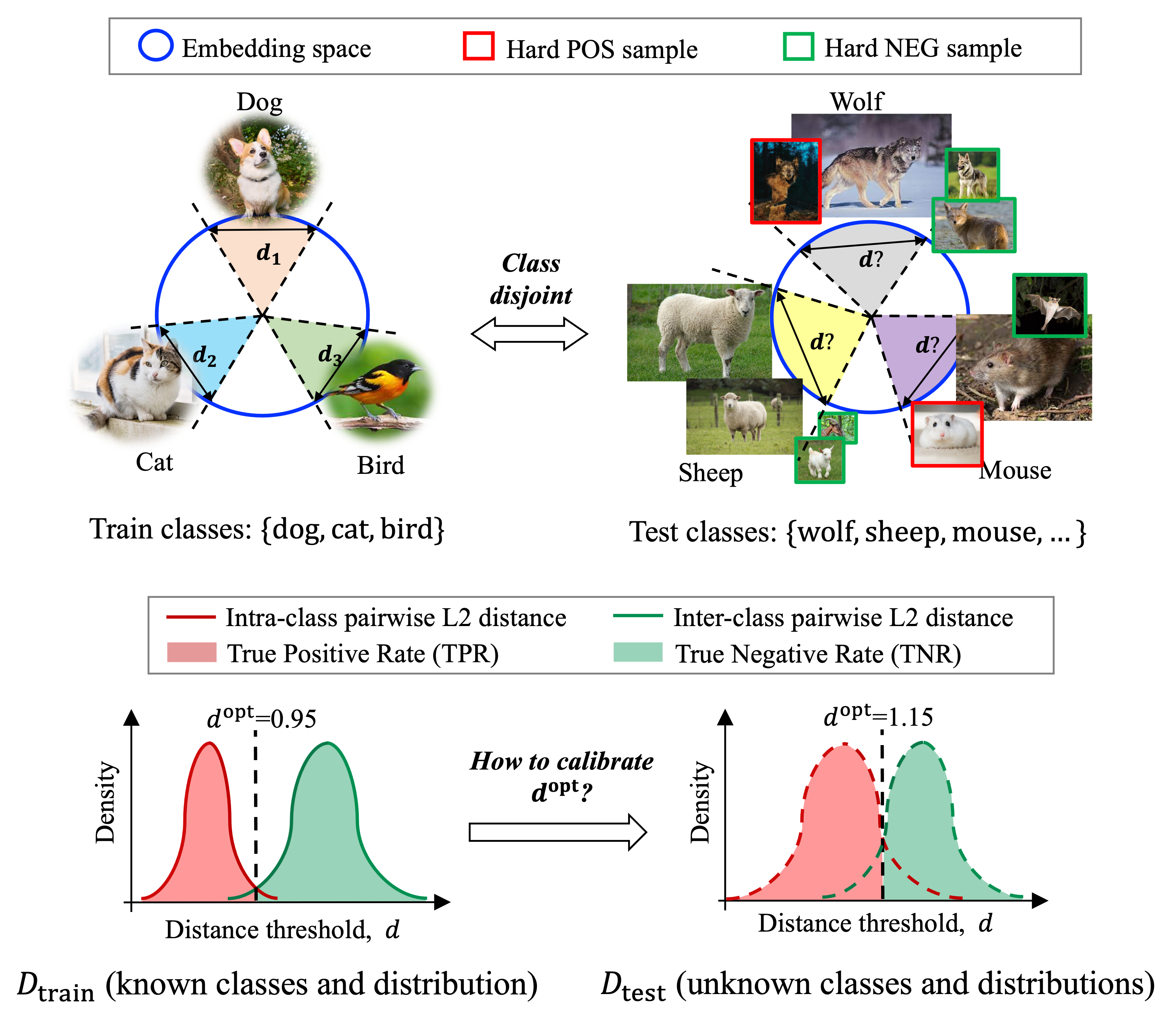}
\vspace{-6pt}
\caption{This figure illustrates the \textbf{\emph{open-world threshold calibration}} problem. In open-world recognition, the embedding model is trained on closed-set classes but tested on distinct open-world classes. When applying the model to open-world classes, it often produces less compact embeddings than those encountered during training,  necessitating the calibration of the distance threshold for achieving the desired TPR and TNR trade-off. However, the absence of prior knowledge about open-world test classes and distributions makes it challenging to find the optimal distance threshold, denoted as $d^\mathrm{opt}$.  
\emph{Best viewed in color}. 
} \label{fig:teaser} 
\vspace{-6pt}
\end{figure}

\section{Introduction}
\label{sec:intro}
In deep metric learning (DML) for visual recognition, distance calibration plays a critical role in determining the user-perceived model performance. Unlike confidence calibration in closed-set classification settings which focuses on aligning confidence probabilities with true likelihood of correctness in a fixed label space~\citep{classification_calibration,mukhoti2020calibrating}, distance calibration in DML aims to pinpoint an optimal distance threshold to achieve a target true positive rate (TPR) or true negative rate (TNR) for diverse test-time distributions~\citep{liu2022oneface}. This calibration is vital because, even with a highly effective embedding model, an inappropriate distance threshold can significantly degrade user experience. The issue becomes more pronounced in open-world recognition, where the embedding model, trained on a closed set of classes (e.g., ${\text{dog, cat, bird}}$), is tested on an open collection of unseen classes (e.g., ${\text{wolf, sheep, mouse, ...}}$). These open-world classes may have very different intra-class and inter-class representation structures, typically being less compact, compared to the training classes. In such open-world scenarios, where prior knowledge about test-time classes and distributions are absent, calibrating the distance threshold becomes a challenging task, as illustrated in \cref{fig:teaser}. We term this task of distance calibration in an open-world scenario as \emph{\textbf{open-world threshold calibration}}. 

Existing posthoc calibration methods, such as~\citep{Platt1999PlattScaling,Zadrozny2001histogram,Zadrozny2002isotonic,Naeini2015Bayesianhistogram,kull2017beta,Guo2017calibration}, typically utilize a fully-labeled calibration dataset that has a similar distribution as the test data~\citep{sugiyama2012machine,ovadia2019can,zhong2021improving} to learn general calibration rules for test distributions. However, this approach has a key limitation: it heavily relies on the assumption of identical distributions between test and calibration data for effective calibration. 
In open-world scenarios, 
this assumption becomes unreliable, posing significant challenges to threshold calibration, including:
\begin{enumerate}[leftmargin=*]
    \item \emph{\textbf{The open-world challenge}} The test data may exclusively contain open-world classes, which exhibit different relationships between distance thresholds and TPR or TNR compared to those encountered during the embedding model training~\cite{liu2022oneface}. Meanwhile, test data composition and quality can vary significantly, potentially exhibiting substantial class imbalances and data corruptions.
    \item \emph{\textbf{Non-stationary data}} In real-world testing environments, the test distribution can be infinitely varied and highly dynamic, rendering the assumption of similar distribution between calibration and test data obsolete. 
    \item \emph{\textbf{Deployment Scalability}} Real-world systems require calibration methods that can adapt to diverse user distributions without individual recalibration. Existing methods lack deployment scalability as they frequently require dedicated calibration data and the creation of specific calibration functions for each user. Imagine a scenario with 1,000 user profiles with distinct classes and data distributions -- creating and deploying custom calibration datasets and functions for each would be impractical. 
\end{enumerate}

Addressing these challenges is crucial for the reliability of DML-based open-world recognition systems. Current posthoc calibration methods are ill-suited for this purpose, as they are inherently inductive and prone to failure when confronted with test data with different distance distributions from the calibration data. 
To address this, we adopt a fresh perspective on distance threshold calibration, treating it as a transductive inference process, where the calibration method incorporates the information of the unlabeled test samples along with the learned calibration rules to make better threshold estimations. 
Our proposed method, \emph{OpenGCN}, employs a Graph Neural Network (GNN), known for its generalization capabilities~\cite{baranwal2021graph,bevilacqua2021size,chuang2022tree,yang2022graph,yang2020learning,hilander,cao2022pss}, to jointly predict pairwise connectivity and two instance-wise representation densities for test data, where the predicted pairwise connectivity is used to compute the TPR and TNR of the test data at each distance threshold to enable transductive threshold calibration. 
OpenGCN is tailored for the task of open-world threshold calibration through 
a carefully crafted learning process, which accurately estimates the mapping between performance metrics and pairwise distance thresholds in open-world scenarios. In particular, the multi-task learning of connectivities and representation densities facilitates information sharing, which 
helps enhance the model's generalization to open-world scenarios~\cite{yang2020learning,hilander}. Additionally, our joint prediction design incorporates two types of density metrics, addressing both intra-class and inter-class connectivity estimations. This approach, as opposed to using a single density metric, is shown to enhance calibration performance, as illustrated in \cref{ablation}. Furthermore, OpenGCN adopts a two-stage training process. It pre-trains on a large closed-world dataset, followed by fine-tuning on a small open-world calibration dataset with disjoint classes to both the closed-world and test data, to adapt the model to be aware of the open-world context. By these design choices, OpenGCN sidesteps the requirement for calibration data to have a similar distance distribution\footnote{We use ``distance distribution" to refer to the distribution of pairwise distances between L2-normalized embeddings from a trained DML model.} as the test data, significantly improving calibration performance in open-world scenarios. 
To summarize, our contributions are as follows:
\begin{enumerate}[leftmargin=*]
\item We are, to the best of our knowledge, the first to formally define the \emph{open-world threshold calibration} problem.
\item We propose \emph{\textbf{T}ransductive \textbf{T}hreshold \textbf{C}alibration (TTC)}, a new threshold calibration paradigm that diverges from traditional inductive posthoc calibration methods, which does not rely on the assumption of similar distance distributions between the test and calibration data.
\item We introduce \emph{OpenGCN}, a GNN-based TTC method tailored for open-world threshold calibration against diverse test distributions. We build comprehensive evaluation protocols with and without distance distribution shifts to assess OpenGCN's performance. The evaluation result underscores OpenGCN's effectiveness and robustness in real-world testing environments. 
\end{enumerate} 

%% file: sec/related_works.tex
\section{Problem Definition and Related Works}\label{problem_def+related_work}
We first introduce some notations and formalize the open-world threshold calibration problem. Let $D_\text{labled}$ be a labeled dataset consisting of two disjoint subsets: $D_\mathrm{train}$ and $D_\mathrm{cal}$, and let $D_\mathrm{test}$ be an unlabeled dataset. In open-world scenarios, the class sets of $D_\mathrm{train}$, $D_\mathrm{cal}$, and $D_\mathrm{test}$, denoted as $C_\mathrm{train}$, $C_\mathrm{cal}$, and $C_\mathrm{test}$, are disjoint, i.e., $C_\mathrm{train}\cap C_\mathrm{cal}=C_\mathrm{train}\cap C_\mathrm{test}=C_\mathrm{cal}\cap C_\mathrm{test}=\varnothing$. 
The goal of open-world threshold calibration is to find a suitable distance threshold that achieves the target TPR and TNR for $D_\mathrm{test}$, given an embedding model trained on $D_\mathrm{train}$.
We approach this as a constrained optimization task, with the objective being maximizing the metric of interest. Take optimizing for TNR with a minimum TPR requirement as an example, this problem can be formulated as follows:
\begin{equation}
\mathop{\mathrm{maximize}}_{d} \mathrm{TNR}_\mathrm{test} \mathrm{, \ \ subject \ to \ } \mathrm{TPR}_\mathrm{test}(d)\geq \alpha \label{eq:optimization}
\end{equation}
where $d$ is the distance threshold, and $\alpha$ is the minimum performance requirement for $\text{TPR}_\text{test}$. Due to the inherent trade-off between TPR and TNR, the objective in \cref{eq:optimization} is equivalent to finding an optimal distance threshold $d^\text{opt}$ for which $\mathrm{TPR}_\mathrm{test}(d^\text{opt})=\alpha$. To solve this, we express $\mathrm{TPR}_\mathrm{test}$ and $\mathrm{TNR}_\mathrm{test}$ at a distance threshold $d$ as follows:
\begin{equation}
\scalemath{0.8}{ 
\mathrm{TPR}_\mathrm{test}(d)={\sum_{i,j\in D_\mathrm{test}}1_{y_i=y_j}\cdot 1_{d_{ij}<d}} \Big/ {\sum_{i,j\in D_\mathrm{test}}1_{y_i=y_j}} 
\label{eq:tpr}}
\end{equation}
\begin{equation}
\scalemath{0.8}{ 
\mathrm{TNR}_\mathrm{test}(d)={\sum_{i,j\in D_\mathrm{test}}1_{y_i\not=y_j}\cdot 1_{d_{ij}>d} } \Big/ {\sum_{i,j\in D_\mathrm{test}}1_{y_i\not=y_j}}}
\label{eq:tnr}
\end{equation}
where $d_{ij}$ is the L2 distance between the embeddings of samples $i$ and $j$, and $y_i$ is the label for sample $i$. The symbol $1_\mathrm{condition}$ represents the indicator function which equals 1 if the condition is met, otherwise 0. With $\text{TPR}_\mathrm{test}$ and $\text{TNR}_\mathrm{test}$ calculated at each distance threshold, we can optimize for the optimal distance threshold $d^\text{opt}$ to achieve the target performance metrics, as described in \cref{eq:optimization}.

\subsection{Related Works}\label{relatedwork}
\noindent\textbf{Open-world Recognition}~\citep{lonij2017open} aims to learn discriminative representations that align distances between representations with their semantic similarities. This allows for effective generalization to diverse, previously unseen open-world classes during testing, setting it apart from closed-set classification where training and testing classes are the same. 
Popular recognition losses~\citep{deng2019arcface,smooth_ap, recall@k_surrogate} typically encourage compact intra-class representations, promoting strong affinity within each class while maintaining separation from other classes. However, it is widely observed that these losses tend to produce highly varied intra-class and inter-class representation structures across classes and distributions~\cite{rippel2015metric,milbich2021characterizing,zhang2024thresholdconsistent}, necessitating threshold calibration to ensure consistent performance across diverse users.

\noindent\textbf{Posthoc Calibration} 
We focus on posthoc calibration methods which are more relevant to our research. Generally, existing posthoc calibration methods fall into two categories: (i) non-parametric methods like isotonic regression~\citep{Zadrozny2002isotonic} and  histogram binning~\citep{Zadrozny2001histogram,Naeini2015Bayesianhistogram}; and (ii) parametric methods such as Platt scaling~\citep{Platt1999PlattScaling} and temperature scaling~\citep{Guo2017calibration}.  
These methods are inductive: they rely on a hold-out calibration set with similar distribution as the test data to derive general rules for fine-tuning the decision threshold, aiming to align the performance metrics with a predefined target. While effective in closed-set classification, these methods struggle in scenarios with significant distribution differences between test and calibration data. Diverging from traditional methods, another group of methods such as conformal prediction~\citep{tibshirani2019conformal,romano2020classification,gibbs2021adaptive,barber2023conformal} or Prediction-Powered Inference~\cite{angelopoulos2023predictionpowered} emphasize confidence coverage guarantees, and has been shown applicable even beyond the setting of exchangeable data~\cite{gibbs2021adaptive,feldman2022conformalized}. However, these methods inherently assume a closed-set setting, making them unsuitable for open-world scenarios. Currently, open-world posthoc calibration remains largely under-explored.

\noindent\textbf{Transductive Inference} Transduction is the reasoning from observed, specific (training) cases to specific (test) cases~\cite{vapnik1999nature}. Such an approach is desirable as it alleviates the problem of overfitting on limited support set since information from the test data is also used for inference. This is also known as increasing VC-dimension for structural risk minimization in classical statistical learning~\cite{joachims1999transductive}. Recently, a large body of works investigated transductive inference for few-shot and open-world recognition tasks~\cite{liu2018learning,qiao2019transductive,hu2020empirical,cao2021open}, where significant increases in performances have been reported. Given the relevance of these tasks, it is worthwhile to reconsider existing inductive posthoc calibration methods for distance threshold calibration in open-world scenarios.

%% file: sec/theory.tex
\begin{figure}[t!]
  \centering
    \includegraphics[width=0.94\linewidth]{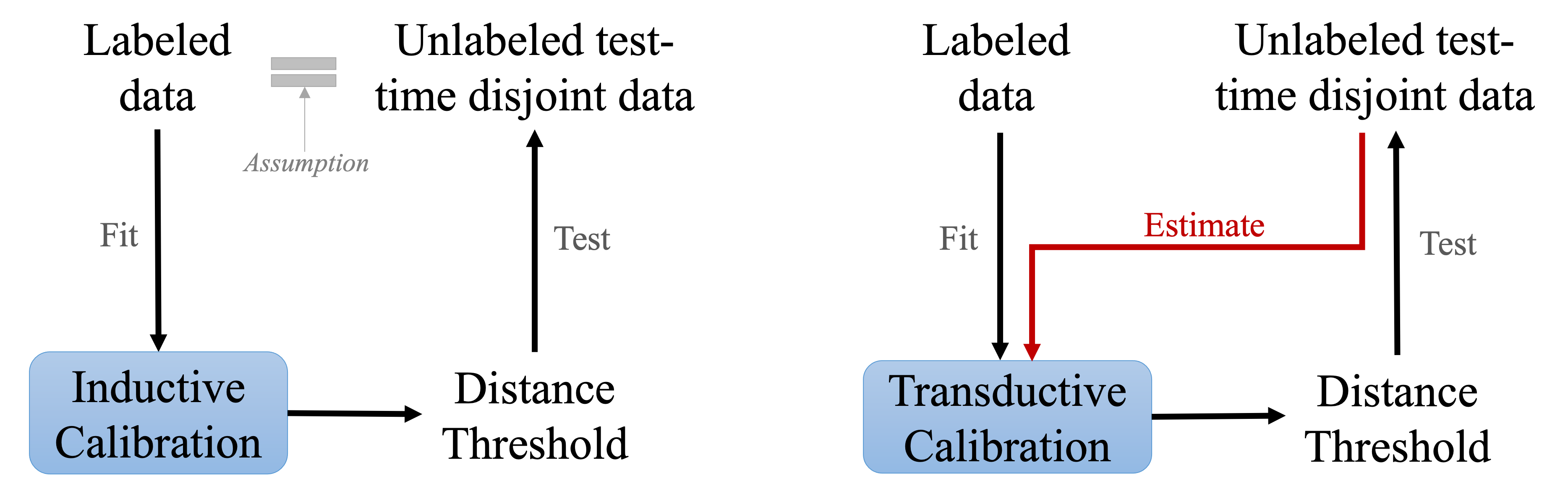}
   \caption{This figure distinguishes between (left) inductive and (right) transductive threshold calibration methods in open-world scenarios with disjoint test-time classes. Inductive methods rely on a labeled hold-out dataset with the same distance distribution as the test data to learn general calibration rules. Transductive methods, however, also use the test information for more specific calibration, as indicated by the {\color{red}red} arrow. 
   \emph{Best viewed in color}.
   } \label{fig:block_diagram}  
\end{figure}

\section{Methodology}\label{methodology}
\subsection{Transductive Threshold Calibration}
Traditional calibration methods are inherently inductive -- they rely on a calibration dataset to learn general calibration rules 
under the assumption of identically distributed data. However, in open-world scenarios, this assumption seldom holds, as the test distribution is unknown and can be infinitely varied and highly dynamic. To improve calibration specificity in the open world, it is natural to adopt a transductive approach, where the TPR and TNR estimations directly involve the test data, rather than relying on a separate calibration dataset that might not accurately represent the test data. As illustrated in \cref{fig:block_diagram}, a transductive approach allows the calibration model to ``see" the unlabeled test data when deciding on the distance threshold, contrasting with the tranditional inductive methods which are ``blind" to the test data. We term this approach as \emph{\textbf{T}ransductive \textbf{T}hreshold \textbf{C}alibration (TTC)}, and the traditional inductive calibration methods as \emph{\textbf{I}nductive \textbf{T}hreshold \textbf{C}alibration (ITC)}.

\begin{figure*}[t!]
  \centering
   \includegraphics[width=0.99\linewidth]{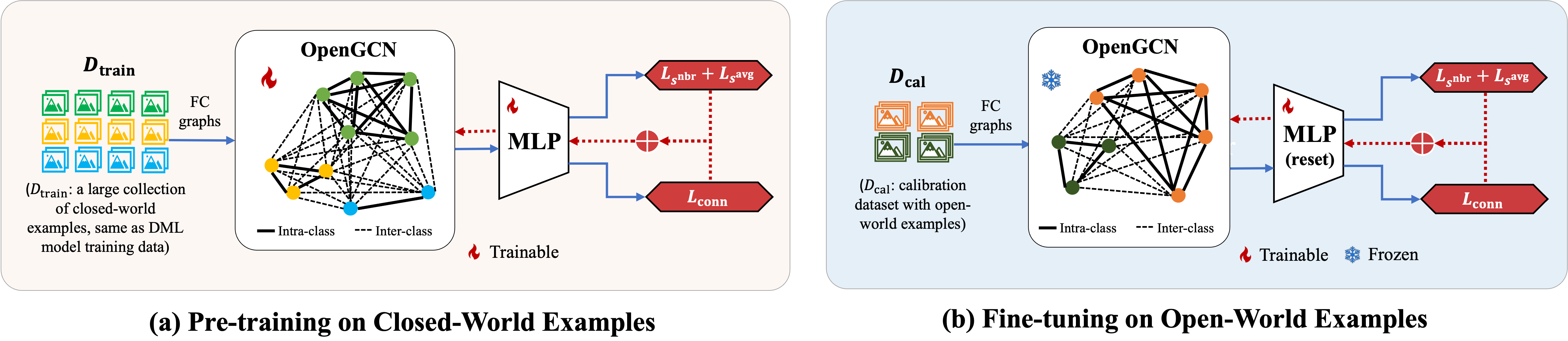}
   \vspace{-1pt}
   \caption{OpenGCN training workflow: (a) During pre-training, OpenGCN jointly optimizes pairwise connectivity, and instance-specific  neighborhood and average densities. (b) During fine-tuning, the 2-layer MLP is reset for fine-tuning, while the other weights remain frozen. Solid {\color{blue}blue} and dashed {\color{red}red} arrows represent forward  and backward propagation, respectively. At test time, we employ the trained OpenGCN model and MLP head to predict the TPR and TNR as functions of each distance threshold specifically for each test distribution. We then follow~\cref{eq:optimization} and use grid search to find the optimal distance threshold for each test dataset. \emph{Best viewed in color.} 
   } \label{fig:Pipeline.} 
\end{figure*}

To overcome the limitations of ITC methods, 
we propose OpenGCN, a GNN-based TTC method with enhanced adaptability and robustness for open-world scenarios with diverse concepts and distance distributions. We highlight the key differences between OpenGCN and conventional ITC methods. First, OpenGCN, as a transductive method, derives distance thresholds by leveraging information directly from the test data. This empowers it to adapt to the characteristics of the test data, thereby eliminating the requirement for the calibration data to share a similar distribution with the test data. 
Second, OpenGCN is engineered to integrate useful information from both closed-world and open-world data sources. This is achieved through a two-stage training process, as illustrated in \cref{fig:Pipeline.}. 
We first pretrain OpenGCN on a closed-world dataset, which is the same dataset used to train the DML embedding model. Afterwards, we fine-tune it on a smaller calibration dataset. This calibration dataset contains open-world classes that do not overlap with those in the test data or the closed-world pretraining data. This approach allows the model to smoothly transition from a closed-world context to open-world scenarios, effectively utilizing closed-world knowledge to enhance its transductive reasoning capabilities in the dynamic and unknown open world. 
In the next section, we delve into the details of OpenGCN, elaborating on how it enables effective TTC for open-world scenarios. 

\subsection{OpenGCN: Learning for Effective TTC}\label{learning_for_OpenGCN}
\noindent\textbf{OpenGCN Inference Workflow} A straight-forward way to estimate $\mathrm{TPR}_\mathrm{test}$ and $\mathrm{TNR}_\mathrm{test}$, as defined in \cref{eq:tpr,eq:tnr}, is to model the true pairwise connectivities with edge connectivity probability~\cite{yang2020learning}. This probability, denoted as $p_{ij}$, quantifies the likelihood that two samples have the same label. By setting a proper connectivity threshold $\tau$, we can approximate $\mathrm{TPR}_\mathrm{test}$ and $\mathrm{TNR}_\mathrm{test}$ as follows:
\begin{equation}
\scalemath{0.825}{ 
\mathrm{\hat{TPR}}_\text{test}(d)={\sum_{i,j\in D_\mathrm{test}}1_{p_{ij}>\tau}\cdot 1_{d_{ij}<d}} \Big/{\sum_{i,j\in D_\mathrm{test}}1_{p_{ij}>\tau}}}
\label{eq:approx-tpr}
\end{equation}
\begin{equation}
\scalemath{0.825}{ 
\mathrm{\hat{TNR}}_\text{test}(d)={\sum_{i,j\in D_\mathrm{test}}1_{p_{ij}\leq\tau}\cdot 1_{d_{ij}>d} } \Big/ {\sum_{i,j\in D_\mathrm{test}}1_{p_{ij}\leq\tau}}}
\label{eq:approx-tnr}
\end{equation}
\indent These formulations offer a TTC solution that centers on precisely predicting pairwise connectivities for open-world test distributions, a problem well-suited for modern deep learning algorithms. Specifically, as shown in \cref{fig:Pipeline.}, OpenGCN is designed as a GNN-based method for predicting pairwise connectivities over graph data constructed from the unlabeled test samples. 
We adopt a GNN architecture, specifically a Graph Attention Network (GAT)~\cite{velivckovic2017graph}, due to its demonstrated effectiveness in generalizing to open-world scenarios~\cite{baranwal2021graph,bevilacqua2021size,chuang2022tree,yang2022graph,yang2020learning,hilander,cao2022pss}. Additionally, we 
use fully connected graphs to ensure that in-graph pairwise distance distribution is representative of the overall pairwise distance distribution. For inference, nodal features extracted by the GAT encoder are concatenated with the original DML embedding features~\cite{yang2020learning} and passed through a 2-layer MLP to predict pairwise connectivities.
The connectivity predictions are then used to transductively estimate TPR and TNR at each distance threshold for the test distributions, following the formulations in \cref{eq:approx-tpr,eq:approx-tnr}, where the connectivity threshold $\tau$ is selected by 10-fold cross validation on $D_\mathrm{cal}$. Due to the typically large size of the test data, for efficient inference, we randomly sample subsets from $D_\mathrm{test}$ to construct fully connected sub-graphs 
for connectivity inference, repeating this process until the TPR and TNR estimations converge. 

\noindent\textbf{Joint Connectivity and Density Estimations} Using representation density prediction as an auxiliary task to enhance connectivity prediction is widely used in clustering tasks~\citep{ankerst1999optics, campello2013density, mcinnes2017hdbscan}. This approach is based on the idea that a cluster typically exists within a contiguous region of high sample density, separated from other clusters. Recent supervised visual clustering works also leverage density as a key modeling parameter to enhance clustering performance by encouraging information sharing between the tasks~\cite{yang2020learning,hilander,yan2021progressive}. Driven by the intrinsic connections between density and connectivity, we adopt a multi-task approach, where we simultaneously learn for pairwise edge connectivity and instance-wise representation densities. However, unlike previous works which only consider one density metric, we simultaneously learn two density metrics: the average density ($s^\mathrm{avg}$), and the neighborhood density ($s^\mathrm{nbr}$). Formally, these two density metrics, defined in~\cite{yang2020learning}, can be expressed as follows\footnote{Although the original definition of $s^\mathrm{avg}$ in \cite{yang2020learning} requires a neighborhood size that includes all samples belonging to a given class, it can be shown by stochastic convergence of random variables that our definition is a tight approximation for \cite{yang2020learning} when $\mathcal{|N}_i|$ is sufficiently large.}:
\begin{equation}
\scalemath{0.825}{ 
s_i^\mathrm{avg}  =\frac{\sum\limits_{j\in\mathcal{N}_i} {a_{ij} \cdot 1_{y_i=y_j}}}{\mathcal{|N}_i|}, \ \ s_i^\mathrm{nbr} =\frac{{\sum\limits_{j\in\mathcal{N}_i} a_{ij} \cdot (1_{y_i=y_j}-1_{y_i\not=y_j})}}{\mathcal{|N}_i|}}
\label{eq:s_avg_s_nbr}
\end{equation}
where $\mathcal{N}_i$ denotes the neighbourhood of a sample $i$, and $a_{ij}$ represents the cosine similarity between the original embedding features of sample $i$ and sample $j$. 

To illustrate the motivation of utilizing both density metrics instead of just one, we first introduce two metrics adapted from prior works~\cite{sra,Directional_Statistics}, namely the class-specific TPR and TNR scores, denoted as $\text{TPR}^k$ and $\text{TNR}^k$, respectively. Let $f_i$ denote the L$_2$-normalized embeddings of an image in a dataset $D$. For a given class $k$, its class-specific TPR and TNR scores can be expressed as:
\begin{equation}
\scalemath{0.825}{ 
\text{TPR}^k = \frac{\|\sum\limits_{i\in D} f_i\cdot 1_{y_i=k}\|}{\sum\limits_{i\in D} 1_{y_i=k}}, \ \ \text{TNR}^k  =\frac{\sum\limits_{i,j\in D} (1-a_{ij}) \cdot 1_{y_j\not=y_i=k}}{\sum\limits_{i,j\in D} 1_{y_j\not=y_i=k}} }
\label{eq:tpr_k_tnr_k}
\end{equation} 
\indent The subsequent theorems formally establish a connection between two density metrics defined in \cref{eq:s_avg_s_nbr} and the class-specific TPR and TNR scores. 

\begin{theorem}\label{theorem1} \textbf{\textup{(Correspondence between \bm{$s^\mathrm{avg}$}} and \bm{$\text{TPR}^k$})} Let $\mathcal{N}$ be a cluster with high purity, where the majority class is $k$. For each sample $i\in\mathcal{N}$, when both $|\mathcal{N}|$ and $|\mathcal{N}_i|$ are sufficiently large, $\text{TPR}^k$ can be approximated as:
\begin{equation}
\scalemath{0.8}{ 
\underset{|\mathcal{N}_i|\rightarrow \infty}{\mathrm{lim}}\text{TPR}^k=\big(\frac{\mathcal{|N}_i|}{2|\mathcal{N}|}\cdot(\underbrace{\underbrace{\frac{1}{|\mathcal{N}|}\sum_{i\in\mathcal{N}}s_i^\mathrm{nbr}}_\mathrm{avg \ s^\mathrm{nbr}}+\underbrace{\frac{1}{|\mathcal{N}|}\sum_{i\in\mathcal{N}}\mathfrak{a_i}^\mathrm{avg}}_{\mathrm{avg} \ \mathfrak{a}^\mathrm{avg}}}_\mathrm{2 \times avg \ s^\mathrm{avg}})\big)^{1/2}\\
\label{eq:theorem1}}
\end{equation}
where $\mathfrak{a}_i^\mathrm{avg}=\frac{1}{|\mathcal{N}_i|}{\sum_{j\in \mathcal{N}_i}a_{ij}}$, and $\mathfrak{a}^\mathrm{avg}$ is the mean of average cosine similarity of all vertices in $\mathcal{N}_i$.
\end{theorem}

\begin{theorem}\label{theorem2} \textbf{\textup{(Correspondence between \bm{$s^\mathrm{avg}-s^\mathrm{nbr}$}} and \bm{$\text{TNR}^k$})} Under the same assumptions in \cref{theorem1}, for a given class $k$, its $\text{TNR}^k$ can be approximated as:
\begin{equation}
\scalemath{0.8}{ 
\underset{|\mathcal{N}_i|\rightarrow \infty}{\mathrm{lim}}\text{TNR}^k = 1-\frac{|\mathcal{N}|}{|\mathcal{N}|_{k^-}}\cdot(\underbrace{\frac{1}{|\mathcal{N}|}\sum_{i\in\mathcal{N}}s_i^\mathrm{avg}-\frac{1}{|\mathcal{N}|}\sum_{i\in\mathcal{N}}s_i^\mathrm{nbr}}_\mathrm{average \ (s^\mathrm{avg} - s^\mathrm{nbr})})}
\end{equation}
where $|\mathcal{N}|_{k^-}$ denotes the number of negative pairs in $\mathcal{N}$ where one sample of each negative pair must have label $k$.
\end{theorem}

Based on the theorems, when the neighborhood size is sufficiently large, considering both density metrics effectively encapsulates both class-specific TPR and TNR within this neighbourhood. As open-world threshold calibration aims to balance the TPR and TNR trade-off for unknown test distributions, it is crucial to capture both aspects to improve within-class and cross-class connectivity predictions. Furthermore, the class-specific nature of these metrics grants them the versatility to adapt to varying class compositions. In \cref{ablation}, we provide an ablation study comparing the use of a single density metric versus both densities, where jointly predicting both densities along with connectivity yields better calibration performance. Thus, we introduce predictions of both density metrics, $s^\mathrm{avg}$ and $s^\mathrm{nbr}$, as auxiliary tasks to enhance the generalization of connectivity prediction. This leads to the following learning objective for training OpenGCN:
\begin{equation}
    \mathcal{L}_\mathrm{overall}=\underbrace{\mathcal{L}_\mathrm{conn}}_\text{main task}+\underbrace{\lambda\cdot(\mathcal{L}_{s^\mathrm{nbr}}+\mathcal{L}_{s^\mathrm{avg}})}_\text{auxiliary task}\label{overall_loss}
\end{equation}
where $\mathcal{L}_\mathrm{conn}$ is the balanced cross-entropy loss for pairwise edge connectivity and $\mathcal{L}_{s^\mathrm{nbr}}$ and $\mathcal{L}_{s^\mathrm{avg}}$ are the mean squared error losses for $s^\mathrm{nbr}$ and $s^\mathrm{avg}$, respectively. Specifically, we define $\mathcal{L}_\mathrm{conn}$ as follows to ensure equal importance for both within-class and cross-class connectivities:
\begin{equation}
\scalemath{0.875}{
\mathcal{L}_\mathrm{conn}=\frac{\sum\limits_{i,j\in V}1_{y_i=y_j}\cdot\mathrm{log}(p_{ij})}{\sum\limits_{i,j\in V}1_{y_i=y_j}}+\frac{\sum\limits_{i,j\in V}1_{y_i\not=y_j}\cdot\mathrm{log}(1-p_{ij})}{\sum\limits_{i,j\in V}1_{y_i\not=y_j}}}
\end{equation}
Meanwhile,  $\mathcal{L}_{s^\mathrm{nbr}}$ and $\mathcal{L}_{s^\mathrm{avg}}$ can be expressed as:
\begin{equation}
\scalemath{0.9}{ 
\mathcal{L}_{s^\mathrm{avg}}=\frac{\sum\limits_{i\in V}(s_i^\mathrm{avg}-\hat{s}_i^\mathrm{avg})^2}{|V|}, \ \mathcal{L}_{s^\mathrm{nbr}}=\frac{\sum\limits_{i\in V}(s_i^\mathrm{nbr}-\hat{s}_i^\mathrm{nbr})^2}{|V|}}
\end{equation}
where $V$ represents the node vertices in the graph data, and $\hat{s}_i$ is the estimated density for each sample based on $p_{ij}$.

\noindent\textbf{Two-stage Training for Adaptability} The DML embedding model, trained on $D_\mathrm{train}$ (closed-set examples), tends to produce more compact embeddings for these examples than those of open-world classes. If OpenGCN is trained solely on $D_\mathrm{train}$, its ability to generalize to the open-world scenarios will be limited. On the other hand, if OpenGCN is trained solely on $D_\mathrm{cal}$, its knowledge may be very narrow since the calibration dataset is typically small and lacks diverse concepts. To tackle this, we borrow established experience in domain generalization and adaptation~\cite{csurka2017domain,wang2018deep,kim2022broad}, and adopt a two-stage training strategy. First, we pretrain OpenGCN on $D_\mathrm{train}$, which consists of a large collection of closed-set examples. After this, we reset the 2-layer MLP while keeping the other parameters frozen. Subsequently, we fine-tune the MLP on $D_\mathrm{cal}$, a small open-world calibration dataset containing disjoint classes to the test data, to adapt the pretrained model to open-world scenarios. In \cref{ablation}, we conduct an ablation study to provide further support for this two-stage training approach. We choose to fine-tune only the MLP based on the practical observations that $D_\mathrm{cal}$ is typically limited in size, and fine-tuning the entire model on such a small dataset may lead to overfitting. It is worth reiterating that this approach does not require additional training data, as the closed-set data is already in place for training the DML embedding model, and the separate open-world calibration dataset is required for conventional inductive posthoc calibration methods as well. 

%% file: sec/experiment_results.tex
\section{Experiment and Result}
We experiment on public recognition benchmarks including iNaturalist-2018~\citep{van2018inaturalist}, CUB-200~\citep{cub} and Cars-196~\citep{cars196}. Below, we outline our setup and present the results. 
Further experiments can be found in the supplementary materials.

\subsection{Dataset and Implementation Details}
\textbf{Datasets} To simulate real-world testing environments
, we consider three calibration scenarios: SameDist, ShiftDist and DiffDist. The SameDist scenario involves cases where $D_\mathrm{cal}$ and $D_\mathrm{test}$ share similar distance distributions, the ShiftDist scenario accounts for test-time non-semantic distance distribution shifts, and the DiffDist 
scenario represents out-of-distribution calibration, where $D_\mathrm{cal}$ and $D_\mathrm{test}$ have very different distance distributions. Note that in all three scenarios, we adhere to the open-world setting where $C_\mathrm{train}\cap C_\mathrm{cal}=C_\mathrm{train}\cap C_\mathrm{test}=C_\mathrm{cal}\cap C_\mathrm{test}=\varnothing$. Below, we elaborate on the setup for each calibration scenario:
\begin{itemize}
\item \emph{\textbf{SameDist}} For iNaturalist, the training and testing classes are distinct, so we directly use the training partition as $D_\mathrm{train}$. To create $D_\mathrm{cal}$, we randomly select 10\% of the test classes, leaving the remaining classes for $D_\mathrm{test}$. For CUB and Cars, where there is overlap between training and testing classes, we divide them into train / cal / test subsets. The train set comprises the first half of the class indices, while the cal / test sets are randomly chosen from the remaining classes with a 1/9 ratio. As $D_\mathrm{cal}$ and $D_\mathrm{test}$ are randomly split from the same dataset, they are expected to have similar distance distributions.
\item \emph{\textbf{ShiftDist}} We consider 13 common image corruption and perturbation types, including noise, blur, weather, and digital distortions, to assess the robustness of the calibration methods under varied adversities. We follow the setups in~\cite{hendrycks2019robustness} and apply the changes to $D_\mathrm{test}$ only, while leaving $D_\mathrm{cal}$ and $D_\mathrm{train}$ unchanged.
\item \emph{\textbf{DiffDist}} To induce significant distance distribution shifts between $D_\mathrm{cal}$ and $D_\mathrm{test}$, we employ the following treatments. For iNaturalist, characterized by a long-tailed distribution, 
we divide its test classes into two sets based on cluster size, each containing approximately the same number of images. For calibration purposes, we use the set with a higher number of images per class (``head" set, denoted as $D_\mathrm{head}$) as $D_\mathrm{cal}$ and the set with fewer images per class (``tail" set, denoted as $D_\mathrm{tail}$) as $D_\mathrm{test}$ to simulate a calibration for the long tail scenario. In addition, we also explore two out-of-domain calibration scenarios. First, for Cars, we transform $D_\mathrm{test}$ into sketches while leaving $D_\mathrm{train}$ and $D_\mathrm{cal}$ untouched. Second, we consider cross-dataset calibration, where the OpenGCN model is pretrained and fine-tuned on iNaturalist (general natural species images) but evaluated on CUB (bird images).
\end{itemize}

\begin{table}[t!]
\caption{Detailed statistics of the datasets.}\label{alldatasets}
\vspace{-3pt}
\centering
\scalebox{0.75}{
\begin{tabular}{ c | c | c | c c c}
\Xhline{2\arrayrulewidth}
\textbf{Setting} & \textbf{Dataset} & \textbf{Partition} & \textbf{\# img} & \textbf{\# cls}  & \textbf{\# img/cls} \\
\Xhline{2\arrayrulewidth}
\multirow{9}{*}{\textbf{SameDist}} & \multirow{3}{*}{Cars} & $D_\mathrm{train}$ & 7,961 & 98 & 81.2\\
&& $D_\mathrm{cal}$ & 866 & 10 & 86.6 \\
&& $D_\mathrm{test}$ & 7,356 & 88 & 83.6\\
\cline{2-6}
&\multirow{3}{*}{CUB} & $D_\mathrm{train}$ & 5,802 & 99 & 58.6 \\
&& $D_\mathrm{cal}$ & 599 & 10 & 59.9\\
&& $D_\mathrm{test}$ & 5,385 & 91& 59.2 \\
\cline{2-6}
& \multirow{3}{*}{iNat} & $D_\mathrm{train}$ & 324,418 & 5,690 & 57.0 \\
&& $D_\mathrm{cal}$  & 12,613 & 245 & 51.5 \\
&& $D_\mathrm{test}$ & 123,047 &  2,207 & 55.8\\
\Xhline{2\arrayrulewidth}
\multirow{1}{*}{\textbf{ShiftDist}} & Cars & \multicolumn{4}{c}{SameDist except for corruption on $D_\mathrm{test}$}\\
\Xhline{2\arrayrulewidth}
\multirow{6}{*}{\textbf{DiffDist}} & \multirow{3}{*}{iNat} & $D_\mathrm{train}$ & \multicolumn{3}{c}{iNat SameDist $D_\mathrm{train}$}\\
&& $D_\mathrm{head}$  & 70,057 & 200 & 350.3 \\
&& $D_\mathrm{tail}$  & 66,036 & 2,252 & 29.3 \\
\cline{2-6}
& Cars & \multicolumn{4}{c}{SameDist except for sketchifying $D_\mathrm{test}$}\\
\cline{2-6}
& \multirow{3}{*}{iNat/CUB} & $D_\mathrm{train}$ & \multicolumn{3}{c}{iNat SameDist $D_\mathrm{train}$}\\
& & $D_\mathrm{cal}$ & \multicolumn{3}{c}{iNat SameDist $D_\mathrm{cal}$} \\
& {\small(cross dataset)} & $D_\mathrm{test}$ & \multicolumn{3}{c}{Entire CUB dataset} \\
\Xhline{2\arrayrulewidth}
\end{tabular}}
\end{table} 

\noindent\textbf{Evaluation Metrics} For a comprehensive evaluation, we consider two approaches to assess calibration performance:
\begin{itemize}
\item \textbf{\emph{Global Evaluation}}: Since we define open-world threshold calibration as the accurate prediction of both TPR and TNR at each distance threshold to meet specific TPR or TNR performance requirements of diverse test-time users, it is natural to employ the combined Mean Absolute Errors (MAE) for both TPR and TNR predictions across the entire distance range as our evaluation metric. Formally, this metric can be expressed as:
\begin{equation}
\begin{split}
\mathrm{MAE}_\text{comb}=\frac{1}{2}\int_{0}^2\big(&|\hat{\text{TPR}}(\mathrm{d})-\text{TPR}(\mathrm{d})|+\\
&|\hat{\text{TNR}}(\mathrm{d})-\text{TNR}(\mathrm{d})| \big) \ d\mathrm{d}
\end{split}
\end{equation}
\item \textbf{\emph{Point-wise Evaluation}}: We first set a performance target and compute the optimal distance threshold, denoted as $\hat{d}^\mathrm{opt}$, based on the TPR or TNR estimations. We then compute the Absolute Error (AE) between the actual performance at $\hat{d}^\mathrm{opt}$ and the target, denoted as $\text{AE}_\text{\text{TPR}}=|\text{TPR}(\hat{d}^\mathrm{opt})-\text{TPR}_\mathrm{target}|$ and $\text{AE}_\text{\text{TNR}}=|\text{TNR}(\hat{d}^\mathrm{opt})-\text{TNR}_\mathrm{target}|$ for TPR and TNR, respectively.
\end{itemize}

\begin{table*}[t!]
\caption{Evaluation in the SameDist scenario using pointwise metrics of $\text{AE}_\mathrm{TPR}$ (optimize for TPR) and $\text{AE}_\mathrm{TNR}$ (optimize for TNR). The smaller the metric, the better. For each dataset, the best and second best results are marked in {\color{red}{Red}} and {\color{blue}{Blue}}, respectively. Shading in the Table: {\color{gray}{Gray}} for posthoc calibration baselines, {\color{cyan}{Cyan}} for clustering baselines, and {\color{blue!50}{Blue}} for our OpenGCN method. \emph{Best viewed in color}. }\label{table:samedist_pointwise}
\vspace{-3pt}
\centering
\scalebox{0.75}{
\begin{tabular}{ c || c c c | c c c  || c c c | c c c || c}
\Xhline{2\arrayrulewidth} 
& \multicolumn{3}{c|}{\textbf{Optimize for TPR=80\%}} & \multicolumn{3}{c||}{\textbf{Optimize for TPR=90\%}} 
& \multicolumn{3}{c|}{\textbf{Optimize for TNR=80\%}} & \multicolumn{3}{c||}{\textbf{Optimize for 
TNR=90\%}}   \\
\cline{2-13}
\textbf{Method} & {Cars} & {CUB} & {Inat}  & {Cars} & {CUB} & {Inat}  & {Cars} & {CUB} & {Inat}  & {Cars} & {CUB} & {Inat} & \textbf{Rank}\\ 
\Xhline{2\arrayrulewidth} 
\rowcolor{gray!10}
Platt scaling~\cite{Platt1999PlattScaling} & 1.35\%& 5.10\%& 6.08\% & {\color{blue}{0.44\%}} & 2.63\% & 4.63\% & 2.83\% & 2.02\% & 7.54\% & 2.93\% & 6.49\% & 0.92\% & 6\\
\rowcolor{gray!10}
Beta calibration~\cite{kull2017beta} & 1.13\%& 5.16\%& 5.51\%& \textbf{{\color{red}{0.02\%}}} & 2.91\% & 3.26\%  & 2.94\% & 1.41\% & 7.57\% & 2.78\% & 6.43\% & 0.93\% & 5\\
\rowcolor{gray!10}
Isotonic regression~\cite{Zadrozny2002isotonic} & {\color{blue}{0.82\%}} & 5.28\% & 4.53\%& 0.90\% & 2.56\% & 3.54\% & 1.94\% & 1.00\% & 5.78\% & 1.26\% & 4.65\% & 0.65\% & 3\\
\rowcolor{gray!10}
Histogram Calibration~\cite{Zadrozny2001histogram} & {\color{blue}{0.82\%}} & 5.28\%& 4.53\%& 0.90\% & 2.56\% & 3.54\% & 1.94\% & 1.00\% & 5.78\% & 1.26\% & 4.65\% & 0.65\% & 4\\
\hline
\rowcolor{LightCyan}
DBSCAN~\cite{ester1996density} & 43.11\% & 18.87\% & \textbf{{\color{red}{0.45\%}}} & 34.57\% & 9.18\% & \textbf{\color{red}{1.85\%}} & 4.09\% & 13.77\% & 12.90\% & 1.60\% & 9.32\% & 9.32\% & 7\\
\rowcolor{LightCyan}
Hi-LANDER~\cite{hilander} & 3.44\%& {\color{blue}{1.36\%}} & 10.54\%& 2.02\% & \textbf{{\color{red}{0.93\%}}} & 7.00\% & \textbf{{\color{red}{0.06\%}}} & {\color{blue}{0.38\%}} & {\color{blue}{2.35\%}} & \textbf{{\color{red}{0.10\%}}} & {\color{blue}{2.20\%}} & {\color{blue}{0.21\%}} & 2\\
\hline
\rowcolor{blue!10}
\textbf{OpenGCN (ours)} &  \textbf{{\color{red}{0.33\%}}} & \textbf{{\color{red}{0.74\%}}} & {\color{blue}{1.59\%}} & 0.72\% &  {\color{blue}{1.41\%}} & {\color{blue}{2.37\%}} & {\color{blue}{0.61\%}} & \textbf{{\color{red}{0.09\%}}} & \textbf{\color{red}{0.74\%}} & {\color{blue}{0.58\%}} & \textbf{\color{red}{0.72\%}} & \textbf{{\color{red}{0.10\%}}} & \textbf{\color{red}{1}}\\
\Xhline{2\arrayrulewidth}
\end{tabular}}
\end{table*}

\noindent\textbf{Baseline Methods} We consider the most representative inductive posthoc calibration methods including Platt Scaling~\cite{Platt1999PlattScaling}, Histogram Calibration~\cite{Zadrozny2001histogram}, Isotonic Calibration~\cite{Zadrozny2002isotonic} and Beta Calibration~\cite{kull2017beta}. Additionally, we explore pseudolabel-based baselines, including traditional clustering methods such as  DBSCAN~\cite{ester1996density} and the state-of-the-art method in GNN-based clustering, Hi-LANDER~\cite{hilander}. For clustering-based methods, we follow their original clustering decoding inference workflows to estimate pseudo labels, and use these pseudo labels to compute $\mathrm{TPR}_\mathrm{test}$ and $\mathrm{TNR}_\mathrm{test}$ for finding $d^\text{opt}$.

\noindent\textbf{Implementation Details} In all experiments, we train ResNet-50 models with 128-dimensional embeddings on $D_\mathrm{train}$ using the setups in~\cite{smooth_ap}. The embedding models are then used to extract the embeddings for $D_\mathrm{train}$, $D_\mathrm{cal}$ and $D_\mathrm{test}$. For training OpenGCN, as implied in \cref{theorem1,theorem2}, the neighborhood size needs to be sufficiently large to encapsulate both intra-class and inter-class representation structures. Thus, we use a batch sizes of 256 for graph construction during training. 
We use the Adam optimizer~\cite{kingma2014adam} with a cosine annealing schedule~\cite{loshchilov2016sgdr}. For traditional calibration methods, we use the official codebase from~\cite{Fabian2020calibrationdetection} to map the ground truth TPR (or TNR) as a function of the distance threshold from $D_\mathrm{train}$ to $D_\mathrm{cal}$. 
When doing point-wise evaluation, the optimal distance threshold $\hat{d}^\mathrm{opt}$ is solved with grid search at a grid size of 0.01. Further details are provided in the supplementary materials.

\subsection{Evaluation Results}
\noindent\textbf{SameDist Calibration} We present the global and pointwise evaluation results for the SameDist scenario in \cref{table:samedist_global} and \cref{table:samedist_pointwise}, respectively. 
For pointwise evaluation, we evaluate at multiple target values (TPR=80\%, 90\% and TNR=80\%, 90\%) to provide a comprehensive assessment. 
Our results reveal that no single calibration method consistently excels across all distance thresholds and datasets. 
However, on average, OpenGCN achieves the highest rank. 
This underscores the importance of TTC in open-world scenarios, where calibration is conducted based on the characteristics of $D_\mathrm{test}$ rather than relying on a calibration dataset that may not accurately represent $D_\mathrm{test}$. 
Additionally, the global metrics in \cref{table:samedist_global} show that, compared to the best baseline method, OpenGCN significantly reduces global error rates by $59.30\%$, $66.49\%$, and $59.15\%$ for Cars, CUB, and iNaturalist, respectively. Among the baseline methods, we observe that DBSCAN performs worse than the traditional posthoc calibration methods, while Hi-LANDER outperforms traditional posthoc methods on Cars and CUB but underperforms on iNaturalist. In contrast, OpenGCN consistently performs well across all three datasets. 

\begin{table}[t!]
\caption{Evaluation in the SameDist scenario using the global error metric of $\mathrm{MAE}_\mathrm{comb}$. 
For each benchmark, the best and second best results are marked in {\color{red}{Red}} and {\color{blue}{Blue}}, respectively. We also report the improvement in error reduction of OpenGCN over the best baseline method. \emph{Best viewed in color}.}\label{table:samedist_global}
\vspace{-3pt}
\centering
\scalebox{0.75}{
\begin{tabular}{ c || c  c c || c}
\Xhline{2\arrayrulewidth} 
\textbf{Method} & \multicolumn{1}{c}{\textbf{Cars}} & \multicolumn{1}{c}{\textbf{CUB}} & \multicolumn{1}{c||}{\textbf{iNat}} & \textbf{Rank} \\ 
\Xhline{2\arrayrulewidth} 
\rowcolor{gray!10}
Platt scaling &  1.55e-2 & 3.59e-2 & 1.23e-2 & 6\\
\rowcolor{gray!10}
Beta calibration &  1.53e-2 &  3.59e-2 & {\color{blue}{1.18e-2}} & 5 \\
\rowcolor{gray!10}
Isotonic regression &  1.38e-2 & 3.61e-2 & {\color{blue}{1.18e-2}} & 3\\
\rowcolor{gray!10}
Histogram calibration  & 1.38e-2 & 3.62e-2 & {\color{blue}{1.18e-2}} & 4 \\
\hline
\rowcolor{LightCyan}
DBSCAN & 1.02e-1 & 1.10e-1 &  3.65e-2 & 7 \\
\rowcolor{LightCyan}
Hi-LANDER &  {\color{blue}{1.29e-2}} & {\color{blue}{1.94e-2}} & 2.14e-2 & 2 \\ 
\hline
\rowcolor{blue!10}
\textbf{OpenGCN (ours)} & \textbf{{\color{red}{5.25e-3}}} & \textbf{{\color{red}{6.50e-3}}} & \textbf{{\color{red}{4.82e-3}}} & \textbf{\color{red}{ 1}}\\ 
\hline
\textbf{Imp. over top baseline} $\uparrow$ & \textbf{59.30\%} & \textbf{66.49\%} & \textbf{59.15\%} & \textbf{69.14\% {\tiny(avg.)}} \\
\Xhline{2\arrayrulewidth}
\end{tabular}}
\end{table}

\begin{table*}[t!]
\caption{Evaluation on the Cars-196 dataset in the ShiftDist scenario across 13 common corruption and perturbation types using combined global error metric of $\mathrm{MAE}_\mathrm{comb}$. The best results are marked in {\color{red}{Red}}. }\label{table:shiftdist_global}
\vspace{-3pt}
\scalebox{0.647}{
\begin{tabular}{ c || c c c | c c c | c c c | c c c c || c }
\Xhline{2\arrayrulewidth} 
& \multicolumn{3}{c|}{\textbf{Noise}} & \multicolumn{3}{c|}{\textbf{Blur}} & \multicolumn{3}{c|}{\textbf{Weather}} & \multicolumn{4}{c||}{\textbf{Digital}} \\
\cline{2-14}
\textbf{Method} & \textbf{Gauss} & \textbf{Shot} &  \textbf{Impulse} & \textbf{Defocus} &
\textbf{Motion} & \textbf{Zoom} & 
\textbf{Snow} & 
\textbf{Fog} & \textbf{Bright} & \textbf{Contrast} & \textbf{Elastic} & \textbf{Pixel} & \textbf{JPEG} & \textbf{Rank} \\ 
\Xhline{2\arrayrulewidth} 
\rowcolor{gray!10}
Platt scaling & 2.95e-2 & 2.99e-2 & 3.41e-2 & 2.66e-2 & 2.62e-2 & 5.02e-2 & 4.24e-2 & 4.37e-2 & 2.16e-2 & 4.61e-2 & 2.16e-2 & 2.24e-2 & 2.03e-2 & 4 \\
\rowcolor{gray!10}
Beta calibration & 2.94e-2 & 2.97e-2 & 3.41e-2 & 2.67e-2 & 2.69e-2 & 5.06e-2 & 4.32e-2 & 4.37e-2 & 2.18e-2 & 4.61e-2 & 2.20e-2 & 2.23e-2 & 2.02e-2  & 5\\ 
\rowcolor{gray!10}
Isotonic regression & 2.88e-2 & 2.85e-2 & 3.38e-2 & 2.37e-2 & 2.31e-2 & 4.85e-2 & 4.07e-2 & 4.34e-2 & 1.83e-2 & 4.59e-2 & 1.85e-2 & 2.03e-2 & 1.80e-2 & 2\\
\rowcolor{gray!10}
Histogram calibration & 2.88e-2 & 2.85e-2 & 3.38e-2 & 2.37e-2 & 2.31e-2 & 4.85e-2 &  4.07e-2 & 4.34e-2 & 1.83e-2 & 4.59e-2 & 1.85e-2 & 2.03e-2 & 1.80e-2 & 3\\
\hline
\rowcolor{LightCyan}
DBSCAN & 4.96e-2 & 6.02e-2 & 7.79e-2 & 9.81e-2 & 1.13e-1 & 1.22e-1 & 1.19e-1 & 4.02e-2 & 9.27e-2 & 4.53e-2 & 1.09e-1 & 1.04e-1 & 8.21e-2 & 7 \\
\rowcolor{LightCyan}
Hi-LANDER & 7.65e-2 & 6.30e-2 & 6.59e-2 & 3.98e-2 & 5.33e-2 &  4.48e-2 & 5.94e-2 & 7.16e-2 & 5.09e-2 & 9.45e-2 & 4.42e-2 & 9.48e-2 & 6.91e-2 & 6\\ 
\hline
\rowcolor{blue!10}
\textbf{OpenGCN (ours)} & \textbf{{\color{red}{1.33e-2}}} & \textbf{\color{red}{5.87e-3}} & \textbf{\color{red}{1.66e-2}} & \textbf{\color{red}{1.50e-2}} & \textbf{\color{red}{1.71e-2}} & \textbf{\color{red}{3.92e-2}} & \textbf{\color{red}{1.42e-2}} & \textbf{\color{red}{7.32e-3}} & \textbf{\color{red}{6.73e-3}} & \textbf{\color{red}{7.08e-3}} & \textbf{\color{red}{5.34e-3}} & \textbf{\color{red}{1.15e-2}} & \textbf{\color{red}{1.68e-2}} & \textbf{\color{red}{1}}\\
\Xhline{2\arrayrulewidth} \textbf{Imp. over top baseline} $\uparrow$ &
\textbf{53.82\%} & 	\textbf{79.40\%}& \textbf{50.89\%}	& \textbf{36.71\%} & \textbf{25.97\%} & \textbf{12.50\%}	& \textbf{65.11\%} & \textbf{81.79\%} & \textbf{63.22\%}	& \textbf{84.37\%}	& \textbf{71.14\%} & \textbf{43.35\%} & \textbf{6.67\%} & \textbf{55.03\% {\tiny(avg.)}} \\
\Xhline{2\arrayrulewidth}
\end{tabular}}
\vspace{-12pt}
\end{table*}
\begin{figure*}[t!]
   \centering
   \includegraphics[width=1\linewidth]{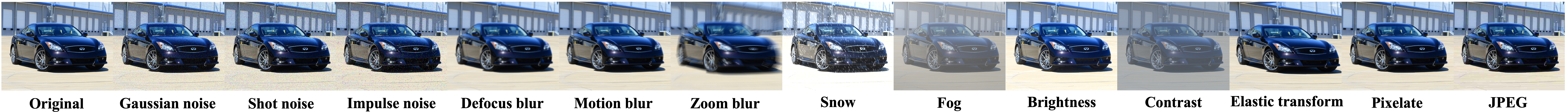}
\end{figure*}

\begin{table}[t!]
\caption{Evaluation in the DiffDist scenario using the global error metric $\text{MAE}_\text{comb}$. The best results are highlighted in {\color{red}{Red}}.}\label{tab:diff_dist}
\vspace{-3pt}
\centering
\scalebox{0.66}{
\begin{tabular}{ c | c c c}
\Xhline{2\arrayrulewidth} 
\textbf{Method} & \textbf{Cars: Sketch} & \textbf{CUB: Cross-dataset} & \textbf{iNat: Longtail} \\
\cline{2-4}
\Xhline{2\arrayrulewidth} 
\rowcolor{gray!10} Platt scaling & 1.08e-1 & 1.15e-1 & 2.09e-2 \\
\rowcolor{gray!10} Beta calibration & 1.08e-1 & 1.15e-1  & 2.12e-2 \\
\rowcolor{gray!10} Isotonic regression & 1.08e-1 & 1.15e-1  & 2.11e-2 \\
\rowcolor{gray!10} Histogram Calibration & 1.08e-1 & 1.15e-1  & 2.11e-2 \\
\hline
\rowcolor{LightCyan} DBSCAN & 5.16e-2 & 1.60e-1 & 7.21e-2 \\
\rowcolor{LightCyan} Hi-LANDER & 6.67e-2 & 1.30e-1 & 6.26e-2 \\
\hline
\rowcolor{blue!10} \textbf{OpenGCN (ours)} & \textbf{\color{red}{3.54e-2}} & \textbf{\color{red}{1.42e-2}} & \textbf{\color{red}{1.82e-2}} \\
\Xhline{2\arrayrulewidth}
\textbf{Imp. over top baseline} $\uparrow$ & \textbf{31.40\%} & \textbf{87.65\%} & \textbf{12.92\%} \\
\Xhline{2\arrayrulewidth}
\end{tabular}}
\end{table}

\noindent\textbf{ShiftDist Calibration} In \cref{table:shiftdist_global}, we report the global error metric $\mathrm{MAE}_\mathrm{comb}$ for each corruption type across various calibration methods. Among the baseline methods, Isotonic Regression and Histogram Calibration appear to be the most effective in the presence of image corruptions. However, it is evident that OpenGCN consistently outperform these baseline methods across all corruption types, achieving an average error reduction of $55.03\%$ compared to the best baseline method. This robust performance against image corruptions can be attributed to the model's pretraining stage, where it was exposed to closed-set data with similar types of corruptions. Additionally, it is observed that, among the various corruption categories, OpenGCN exhibits the most improvement in the weather category, while showing the least improvement in the blur category.

\noindent\textbf{DiffDist Calibration} We present the DiffDist calibration results in \cref{tab:diff_dist}. As observed, in this scenario characterized by a substantial shift in distance distributions between $D_\mathrm{cal}$ and $D_\mathrm{test}$, all calibration methods display elevated errors compared to the SameDist scenario. However, OpenGCN demonstrates superior performance compared to the other calibration methods in both out-of-domain settings (sketch and cross-dataset) and the long-tail calibration setting, achieving an average relative reduction in the global error $\mathrm{MAE}_\mathrm{comb}$ of 43.99\%. In particular, we observe significant improvement in the cross-dataset setting (pretrained and finetuned on the iNaturalist-2018 nature species dataset but tested on the CUB birds dataset), where OpenGCN achieves a notable error reduction of 87.65\%.

\begin{table}[t!]
\caption{Impact of multi-task learning on global error metric $\text{MAE}_\text{comb}$ on iNaturalist-2018. We use $\lambda=10$ for all experiments.}\label{table:multitask}
\vspace{-3pt}
\centering
\scalebox{0.7}{
\begin{tabular}{ c | c | c  c  c c} 
\Xhline{2\arrayrulewidth} 
& \textbf{Best}
& \multicolumn{4}{c}{\textbf{OpenGCN loss ablations}} \\
\cline{3-6}
& \textbf{baseline} & $\mathcal{L}_\text{conn}$ & $+\lambda\cdot\mathcal{L}_{s^\text{avg}}$ & $+\lambda\cdot\mathcal{L}_{s^\text{nbr}}$ & $+\lambda\cdot(\mathcal{L}_{s^\text{avg}}+\mathcal{L}_{s^\text{nbr}})$ \\ 
\cline{2-4}
\Xhline{2\arrayrulewidth} 
$\text{MAE}_\text{comb}$ & 1.18e-2 & 6.25e-3 & 5.37e-3 & 5.12e-3 & \textbf{4.82e-3} \\
\Xhline{2\arrayrulewidth}
\end{tabular}}
\end{table}

\begin{table}[t!]
\caption{Impact of fine-tuning on open-world calibration dataset 
on global error metric $\text{MAE}_\text{comb}$. PT: pretraining FT: finetuning. Numbers in the bracket show the relative improvement over PT.}\label{table:ablation_two_stage}
\vspace{-3pt}
\centering
\scalebox{0.73}{
\begin{tabular}{ c | c  c  c} 
\Xhline{2\arrayrulewidth} 
\textbf{Method} & \multicolumn{1}{c}{\textbf{Cars}} & \multicolumn{1}{c}{\textbf{CUB}} & \multicolumn{1}{c}{\textbf{iNat}}  \\ 
\cline{2-4}
\Xhline{2\arrayrulewidth} 
OpenGCN (PT) & 2.90e-2 & 2.52e-2 & 3.55e-2 \\
OpenGCN (PT+FT) & 5.25e-3 \textbf{{\small{(81.9\%)}}} & 6.50e-3 \textbf{{\small{(74.2\%)}}} & 4.82e-3 \textbf{{\small{(86.4\%)}}} \\
\Xhline{2\arrayrulewidth}
\end{tabular}}
\end{table}

\subsection{Ablation Studies}\label{ablation}
\noindent\textbf{Importance of Multi-task Learning} We assess the impact of multi-task learning on $\text{MAE}_\text{comb}$. As shown in \cref{table:multitask}, compared to predicting connectivity only, employing a single density metric in conjunction with connectivity prediction helps reduce $\text{MAE}_\text{comb}$ from 6.25e-3 to 5.37e-3 for $s^\text{avg}$ and to 5.12e-3 for $s^\text{nbr}$, respectively. However, by utilizing both density metrics, we further decrease this error to 4.82e-3. This supports our choice to incorporate both density metrics, allowing us to capture both intra-class compactness and inter-class separation while facilitating information sharing for improved connectivity prediction.

\noindent\textbf{Importance of Two-stage Training} We assess the impact of two-stage training on OpenGCN by comparing $\text{MAE}_\text{comb}$ before and after fine-tuning on $D_\text{cal}$ across all three benchmarks. The comparison in \cref{table:ablation_two_stage} reveals significant error reduction of up to 86.4\% after fine-tuning on the open-world calibration dataset. This results supports our choice of two-stage training in adapting the calibration model from the closed-world context to the open-world scenarios. 

%% file: sec/conclusions.tex
\section{Conclusions}
In this work, we formally define the open-world threshold calibration problem for DML-based open-world visual recognition systems. To address this problem, we introduce OpenGCN, a GNN-based transductive threshold calibration method designed to enhance adaptability in open-world scenarios. Unlike traditional posthoc calibration methods, OpenGCN does not rely on the common assumption of matching distance distributions between $D_\mathrm{cal}$ and $D_\mathrm{test}$. Instead, it leverages the information of the unlabeled test instances along with learnt calibration rules to predict pairwise connectivity of the test data, via a GNN, to enable effective transductive threshold calibration in open-world scenarios. Our evaluations demonstrate that OpenGCN outperforms both traditional posthoc calibration methods and pseudolabel-based calibration techniques. When assessed using global error metrics, OpenGCN exhibits significant improvements, achieving average error reductions of $69.14\%$, $40.85\%$, and $22.58\%$ for SameDist, ShiftDist, and DiffDist calibration scenarios, respectively, compared to the best baseline method. Overall, our results underscore OpenGCN's robustness across different distance distribution patterns between $D_\mathrm{cal}$ and $D_\mathrm{test}$, highlighting its practical applicability for threshold calibration in DML-based open-world recognition applications.

\noindent\textbf{Limitations} OpenGCN is computationally less efficient and more susceptible to over-parameterization compared to traditional posthoc calibration methods. 
Furthermore, OpenGCN is not a calibration-data-free method as it still requires some calibration data in addition to the closed-world data used for training the embedding model. 